\def\paperlanguage{}
\pdfoutput=1 

\documentclass[letterpaper, 10 pt, conference]{ieeeconf}  

\usepackage{bm}
\usepackage{cite}
\usepackage{flushend}
\include{preamble}

\newcommand{\ctext}[1]{\raise0.2ex\hbox{\textcircled{\scriptsize{#1}}}}

\IEEEoverridecommandlockouts                              

\title{\LARGE \textbf
  {
    \switchlanguage%
    {%
      Daily Assistive Modular Robot Design Based on\\Multi-Objective Black-Box Optimization
    }%
    {%
      多目的ブラックボックス最適化に基づく生活支援モジュラーロボット設計
    }%
  }
}

\author{Kento Kawaharazuka$^{1}$, Tasuku Makabe$^{1}$, Kei Okada$^{1}$, and Masayuki Inaba$^{1}$
  \thanks{$^{1}$ The authors are with the Department of Mechano-Informatics, Graduate School of Information Science and Technology, The University of Tokyo, 7-3-1 Hongo, Bunkyo-ku, Tokyo, 113-8656, Japan.
    {\texttt\small [kawaharazuka, makabe, k-okada, inaba]@jsk.t.u-tokyo.ac.jp}
  }
}
\begin{document}

\maketitle
\thispagestyle{empty}
\pagestyle{empty}

\begin{abstract}
  \switchlanguage%
  {%
    The range of robot activities is expanding from industries with fixed environments to diverse and changing environments, such as nursing care support and daily life support.
    In particular, autonomous construction of robots that are personalized for each user and task is required.
    Therefore, we develop an actuator module that can be reconfigured to various link configurations, can carry heavy objects using a locking mechanism, and can be easily operated by human teaching using a releasing mechanism.
    Given multiple target coordinates, a modular robot configuration that satisfies these coordinates and minimizes the required torque is automatically generated by Tree-structured Parzen Estimator (TPE), a type of black-box optimization.
    Based on the obtained results, we show that the robot can be reconfigured to perform various functions such as moving monitors and lights, serving food, and so on.
  }%
  {%
    ロボットの活動範囲は, 環境の固定された産業から介護支援や日常生活支援等, 多様で変化する環境へと広がりつつある.
    特に, ユーザやタスクごとにパーソナライズされたロボットの自律的な構築が求められている.
    この問題に対し我々は, 多様なリンク構成への組み替えが可能かつ, ロックして重いものを運んだり逆にフリーにして手で容易に動かしたりすることが可能なアクチュエータモジュールを開発している.
    所望の動作点を複数与え, これを満たし必要トルクが最小なモジュラーロボット構成を, ブラックボックス最適化の一種であるTree-structured Parzen Estimator (TPE)により自動生成する.
    得られた結果に基づきロボットを組み替えることで, モニタやライトの移動, 配膳等の多様な機能をロボットに付与できることを示す.
  }%
\end{abstract}

\section{INTRODUCTION}\label{sec:introduction}
\switchlanguage%
{%
  Robots are expanding their field of activities from industrial fields, where the environment is fixed, to diverse and changeable environments including nursing care support and daily life support \cite{okada2005daily, saito2011subwaydemo}.
  In this context, instead of introducing a large number of robots that are exactly the same, robots that can be personalized for each user and task are needed.
  When such robots perform daily assistive tasks, e.g. move monitors, whiteboards, fans, and lights, serve food, and position tables for smaller robots, the actions vary greatly depending on the user, environment, and task.
  Of course, a general-purpose six-axis arm robot can be used, but if the robot can be configured for a given task with fewer joints and minimized torque requirements, the task can be carried out more continuously and efficiently.
  In addition to actually performing a task, the robot can be used as a table or chair by locking its posture when it is not moving, or can be manually operated by leaving its joints released.

  Therefore, we propose an actuator module with a lock-release mechanism that can be reconfigured to various link configurations, and a robot design optimization method based on this module.
  By freely reconfiguring this module, the robot can perform personalized movements with a small number of appropriate joints.
  Also, by using the lock-release mechanism the robot can carry heavy objects and be operated by direct teaching.
  Here, multiple target coordinates or trajectories for a task are given, and a robot configuration that satisfies both the minimization of the control errors and the minimization of the required torque is automatically generated.
  Many robot models are generated by randomizing joint module types, joint orientations, and link lengths, and these parameters are optimized by using multi-objective black-box optimization.
  We analyze the robot configurations obtained under various conditions and show that the desired task can actually be performed by reconfiguring the actuator modules.
}%
{%
  ロボットは環境が固定された産業分野から, 介護支援や日常生活支援を含む, 多様で変化しうる環境へと, その活躍の場を広げつつある\cite{okada2005daily, saito2011subwaydemo}.
  それに伴い, 全く同じ形のロボットを大量に導入する形から, ユーザやタスクごとにパーソナライズされる形のロボットが必要となっている.
  例えば, モニタやホワイトボード, 扇風機, ライトを動かしたり, 配膳やより小さなロボットのための台の位置決めを行ったりと, 多様な動作があり, それはユーザや環境, タスクごとに大きく異なる.
  もちろん汎用的な6軸アームを用いる方法もあるが, 指定のタスクに対してより少ない関節数かつ必要トルクを最小化した形でロボットを構成できれば, より継続的かつ有利にタスクを進めることができる.
  また, ロボットが実際にタスクをこなすだけでなく, 動かないときは姿勢をロックして台として用いたり, 関節をフリーにして人間が移動せたり教示を行ったりする場合もある.

  この課題に対して, 本研究では多様なリンク構成への組み替えが可能かつ, ロック・フリー機構を有するアクチュエータモジュールと, これ基づくロボット設計最適化を提案する.
  このモジュールを自由に組み替えることで, パーソナライズされた動作を少ない適切な関節数で行い, かつ重量物の運搬や人間による教示を可能とする.
  この際, タスクにおける所望の動作点や軌道を複数与え, それらの実現誤差の最小化と, 必要トルクの最小化の2点を満たすロボット構成を自動生成する.
  多目的のブラックボックス最適化を用いて, 関節モジュールの種類や関節の向き, リンク長をランダマイズしたモデルを自動生成, それらパラメータの最適化を行う.
  多様な条件で得られたロボット構成について解析し, 実際にアクチュエータモジュールを組み替えて, 所望のタスクが実行できることを示す.
}%

\begin{figure}[t]
  \centering
  \includegraphics[width=0.95\columnwidth]{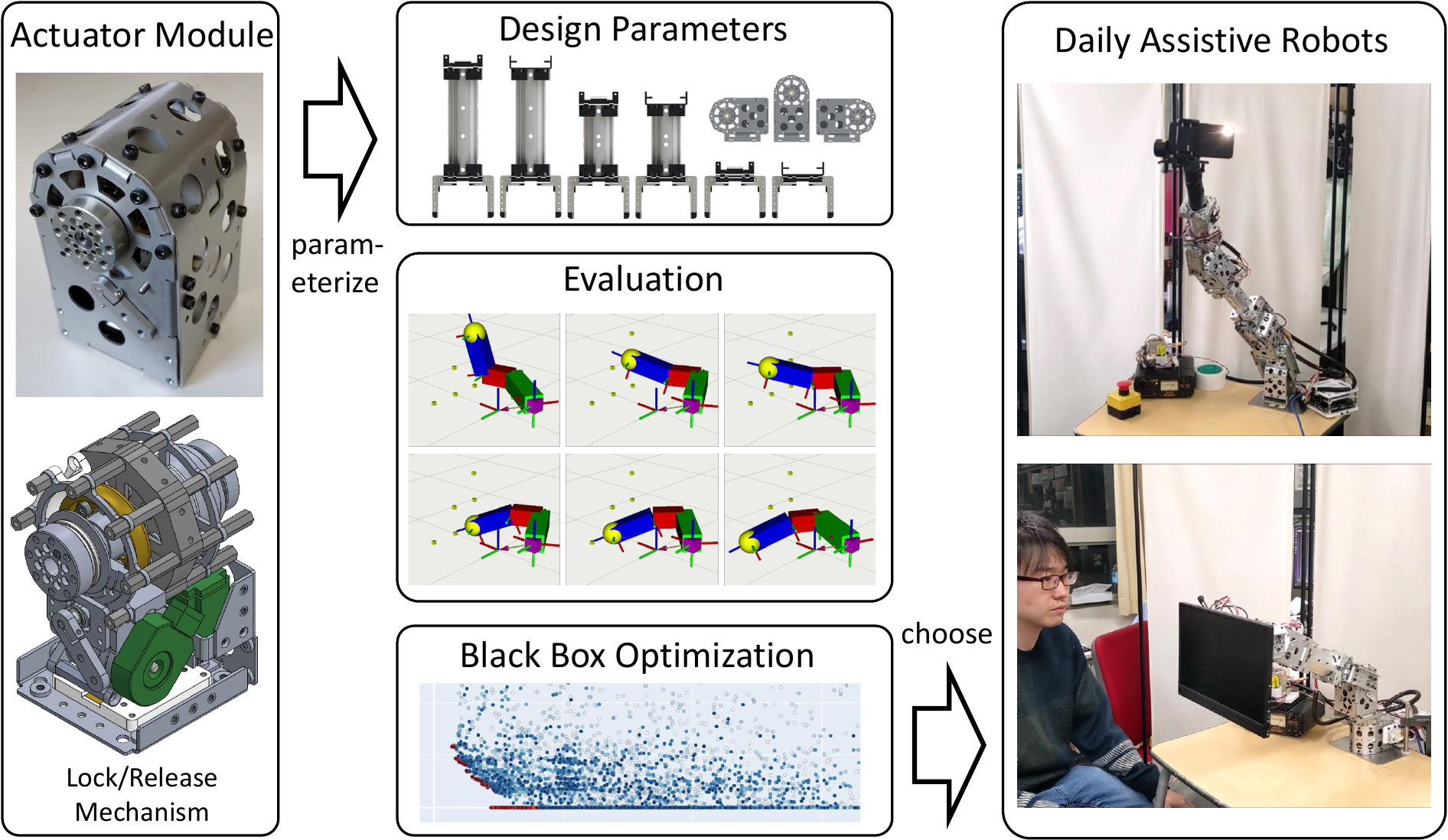}
  \caption{The concept of this study: automatic design optimization of robots with actuator modules with a lock-release mechanism through multi-objective black-box optimization.}
  \label{figure:concept}
\end{figure}

\begin{figure*}[t]
  \centering
  \includegraphics[width=1.95\columnwidth]{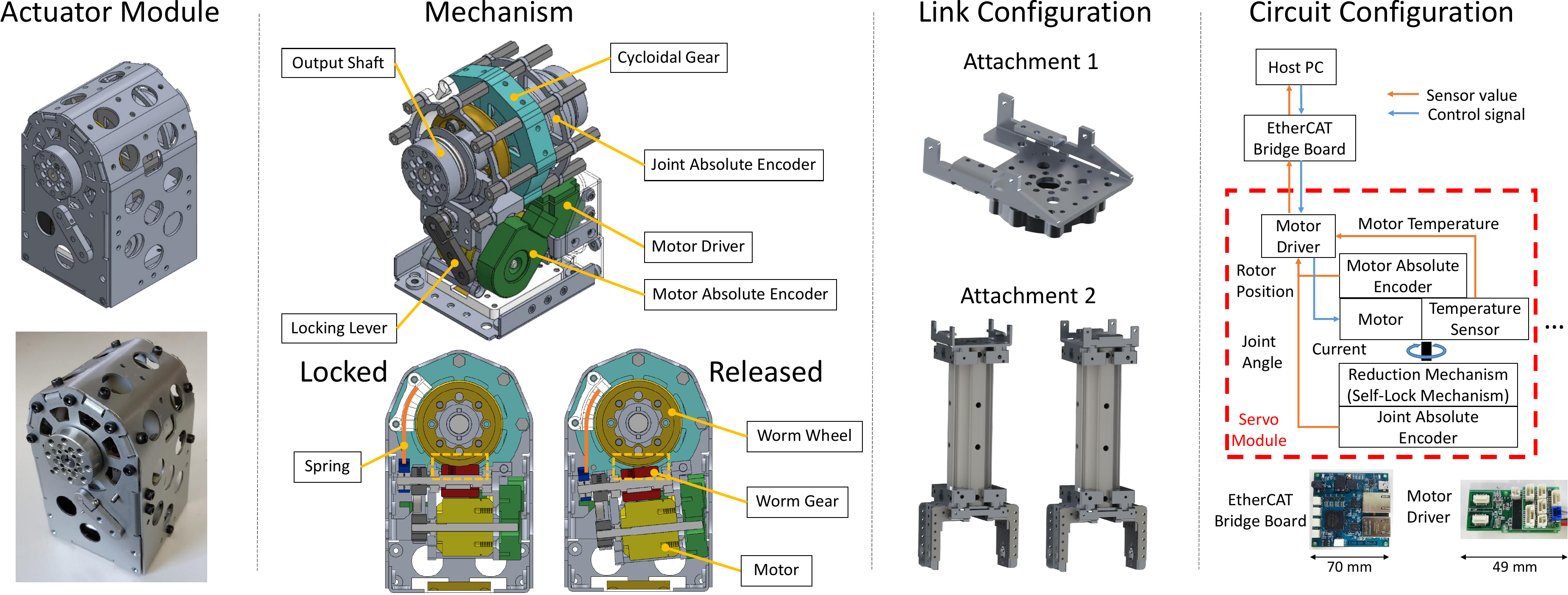}
  \caption{The modular actuator design with a lock-release mechanism. This module includes a worm gear and cycloidal gear that can switch between locked and released states by a locking lever. By using two joint attachments, various kinematic configurations can be generated.}
  \label{figure:module}
\end{figure*}

\switchlanguage%
{%
  Modular robot design and its design optimization have been studied extensively.
  For industrial robots, the number and type of modules and the relative positions among the modules satisfying the target coordinates are optimized by a genetic algorithm \cite{yang2000modular}.
  In \cite{xiao2016designopt}, though not a modular-type robot, the motor type and gear ratio of a general six-axis manipulator are optimized based on the minimization of weight and the maximization of manipulability.
  \cite{zhao2020robogrammar} has optimized the joint arrangements and link lengths of a modular robot that can run on uneven terrain.
  \cite{hu2022modular} has performed design optimization of a modular robot that can run on uneven terrain using Generative Adversarial Network (GAN) \cite{goodfellow2014gan}.
  On the other hand, previous studies are not conducted on actual robots \cite{yang2000modular, xiao2016designopt, zhao2020robogrammar, hu2022modular}, do not perform optimization for continuous values but only for discrete values \cite{yang2000modular}, do not vary the number of modules \cite{xiao2016designopt}, or do not perform multi-objective optimization \cite{yang2000modular, hu2022modular}.
  In addition, there are few examples of modular robots that include a lock-release mechanism for the purpose of personalization in daily assistive robots.
  In this study, various Unified Robot Description Format (URDF) models are automatically generated from the defined constraints, the feasibility of target coordinates and necessary torques are evaluated, and the design parameters are optimized by using Tree-Structured Parzen Estimator (TPE) \cite{bergstra2011tpe}, a type of black-box optimization.

  The structure of this study is as follows.
  In \secref{sec:design}, we describe an actuator module which can be reconfigured to various link configurations and has a lock-release mechanism, as well as other link structures and circuit configurations.
  In \secref{sec:opt}, a black-box optimization procedure is presented to automatically output the link configuration that can satisfy the target coordinates with minimum torque.
  In \secref{sec:experiment}, the link structure obtained by the optimization is analyzed in simulation, and the actual robot experiments with the developed actuator modules are conducted.
  In \secref{sec:discussion}, we discuss the experimental results and some limitations of this study, and conclude in \secref{sec:conclusion}.
}%
{%
  これまでもモジュラーロボット設計とその設計最適化について研究が行われてきた.
  \cite{yang2000modular}は産業用ロボットに向け, 所望動作点を満たすモジュールの数と種類, モジュール間の相対的な位置を遺伝的アルゴリズムにより最適化した.
  \cite{xiao2016designopt}はモジュールではないが, 重量の最小化とmanipulabilityの最大化に基づき一般的な6自由度マニピュレータのモータとギア比を最適化した.
  \cite{zhao2020robogrammar}は不整地を走行可能なモジュールロボットの関節配置やリンク長に関する最適化を行った.
  \cite{hu2022modular}はGenerative Adversarial Network (GAN)を用いた多様な身体生成と不整地走行可能なモジュールロボットの設計最適化を行った.
  一方で, 多くの例は実際のロボット構成を行っていない\cite{yang2000modular, xiao2016designopt, zhao2020robogrammar, hu2022modular}, 離散値のみで連続値の最適化を行わない\cite{yang2000modular}, モジュール数が変化しない\cite{xiao2016designopt}, 多目的な最適化を行いわない\cite{yang2000modular, hu2022modular}等の問題がある.
  また, 介護支援や日常生活支援におけるパーソナライズを目的としてフリーやロック機構を含み構築されたモジュラーロボットの例はほとんど存在しない.
  本研究は定めた制約から大量のUnified Robot Description Format (URDF)を自動生成, 所望動作点の実現度と必要トルクを評価し, これをブラックボックス最適化の一種であるTree-Structured Parzen Estimator (TPE) \cite{bergstra2011tpe}によって多目的最適化している.

  本研究の構成は以下である.
  \secref{sec:design}では, 多様なリンク構成への組み替えが可能かつロックとフリー機構を有するアクチュエータモジュール, その他リンク構造や回路構成について述べる.
  \secref{sec:opt}では, 所望の動作点を最小のトルクで満たすことが可能なリンク構成を自動出力するブラックボックス最適化の手順について示す.
  \secref{sec:experiment}では, 最適化により得られたリンク構造の解析と, アクチュエータモジュールを使った実際のロボット構成の実現について実験を行う.
  \secref{sec:discussion}では, 本研究における実験結果といくつかの限界について考察し, \secref{sec:conclusion}で結論を述べる.
}%

\section{Modular Actuator Design with Lock-Release Mechanism} \label{sec:design}
\switchlanguage%
{%
  The configuration of the actuator module in this study is shown in \figref{figure:module}.
  This is a 0.07$\times$0.07$\times$0.115 [m] actuator module with a square bottom.
  The motor inside is a T-MOTOR MN2212, and a worm gear and cycloid gear are used.
  The gear ratio of the worm gear is fixed at 50:1, which satisfies the self-lock condition.
  The gear ratio of the cycloid gear is variable, but basically set to 47.
  The actuator module has a lock-release mechanism.
  The worm gear and the worm wheel can be disengaged by rotating a locking lever, and the state can be switched between lock (the motor is connected to the output shaft) and release.
  The locked state enables the robot to carry heavy objects without back drive, and the released state enables the robot to be operated by direct teaching.

  The link configuration is described below.
  Two types of attachments can be attached to one or both ends of the rotation axis of the actuator module.
  The bottom of the actuator module and the attachments, and their screws are arranged in a square shape, so that various connections can be realized.
  Although the link length of Attachment 2 of \figref{figure:module} can be changed continuously, it is handled as a discrete value in this study for the sake of practicality.

  The circuit configuration is described below.
  HOST PC can be daisy-chained to Motor Driver through EtherCAT Bridge Board.
  The motor driver receives information from a temperature sensor, motor-side absolute encoder, and joint-side absolute encoder, and calculates motor current commands.
  Since the motor drivers are very small and each of them is equipped with an IMU, they can be used as redundant sensors.
  This compact module includes a motor, reduction gears, encoders on the motor and joint side, a temperature sensor, an inertial sensor, and a motor driver.
  The maximum current for the motor driver is 10A.
}%
{%
  本研究のActuator Moduleの構成を\figref{figure:module}に示す.
  これは0.07$\times$0.07$\times$0.115 [m]のアクチュエータモジュールであり, 底面が正方形の形をしている.
  モータはT-MOTOR MN2212を用いており, ギアはウォームギアとサイクロイド減速機を使用している.
  ウォームギアのギア比は固定で50:1であり, 逆駆動しない条件を満たす.
  サイクロイド減速機のギア比は可変で, 基本は47を用いている.
  アクチュエータモジュールはロック・フリー機構を有する.
  ロッキングレバーを回転させることで, ウォームギアとウォームホイールの噛み合わせ外すことができ, ロック状態とフリー状態の切替が可能である.
  ゆえに, ロックすることで逆駆動せずに重い物体を運んだり, フリーにして動作教示等が可能となっている.

  リンクの構成について述べる.
  2種類のアタッチメントがActuator Moduleの片端または両端にそれぞれ取り付け可能である.
  Actuator Moduleとアタッチメントの底辺, それらのネジの間隔が正方形をしているため, 多様な接続を実現可能である.
  なお, \figref{figure:module}のAttachment 2はリンク長を連続的に変化可能であるが, 本研究では実用性の観点から離散値として扱っている.

  回路構成について述べる.
  HOST PCはEtherCAT Bridge Boardを通してMotor Driverにデイジーチェーンで接続することができる.
  モータドライバは温度センサ, モータ側アブソリュートエンコーダ, 関節側アブソリュートエンコーダの情報を受け取り, 電流指令をモータに送る.
  モータドライバは非常に小型であり, それぞれIMUを搭載するため, これを冗長センサとして利用することも可能である.
  モータ, 減速機, モータやジョイント側のエンコーダ, 温度センサ, 慣性センサ, モータドライバを含んだコンパクトなモジュールとなっている.
  なお, 使用するモータドライバの最大電流は10Aとなっている.
}%

\begin{figure*}[t]
  \centering
  \includegraphics[width=1.8\columnwidth]{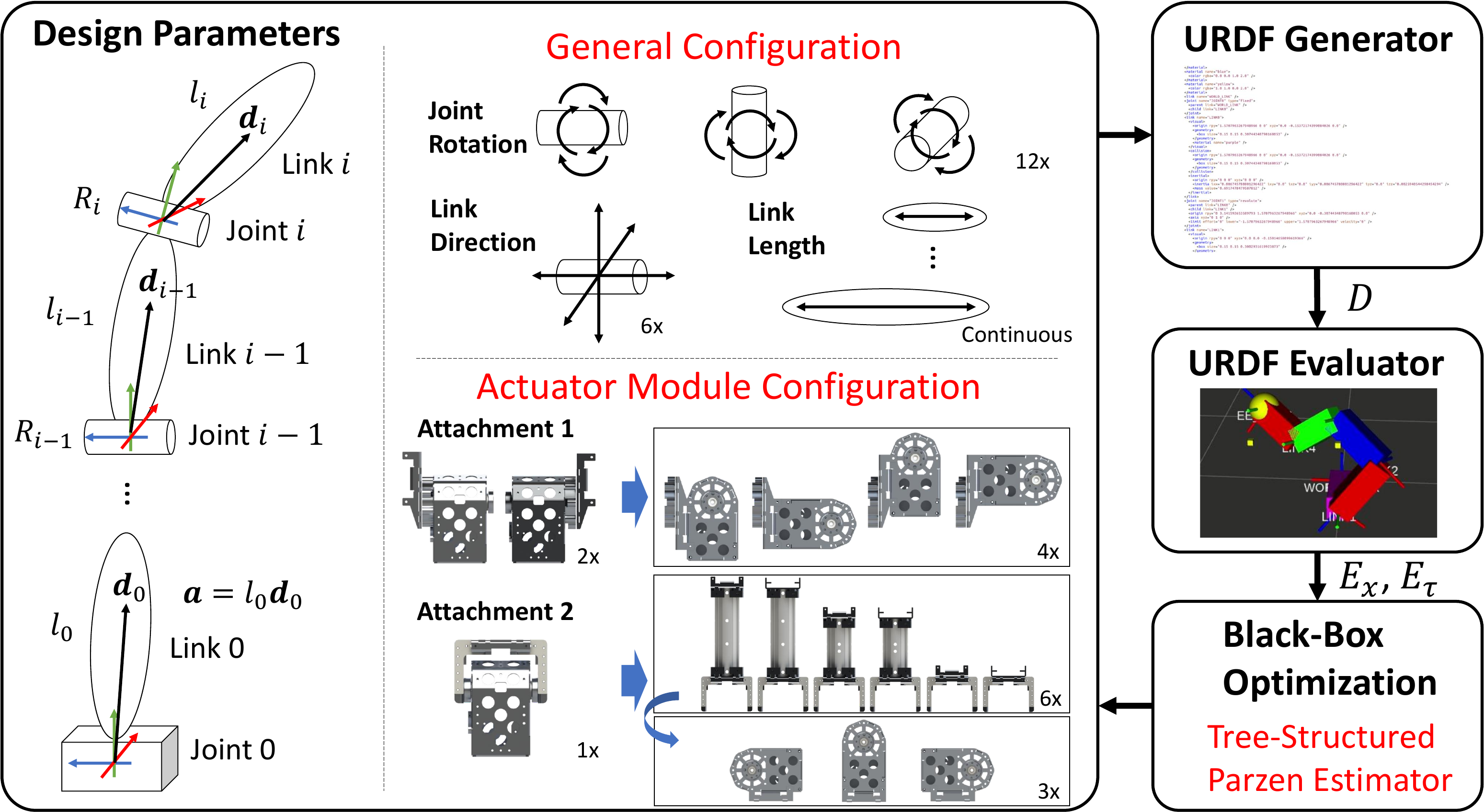}
  \caption{The automatic robot design system with black-box optimization. We prepare different design parameters for general configuration and actuator module configuration, generate a URDF model automatically, evaluate it, and optimize the design parameters through Tree-Structured Parzen Estimator.}
  \label{figure:system}
\end{figure*}

\begin{figure}[t]
  \centering
  \includegraphics[width=0.95\columnwidth]{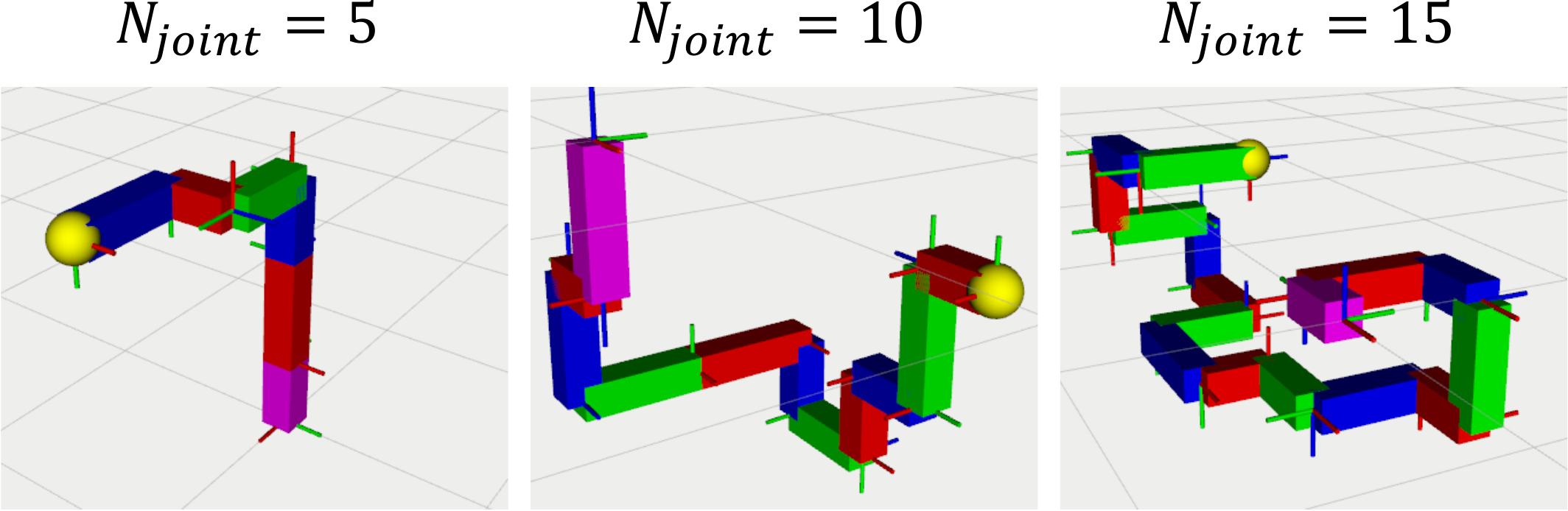}
  \caption{Examples of robots generated with general configuration.}
  \label{figure:general-example}
\end{figure}

\section{Automatic Robot Design with Multi-Objective Black-Box Optimization} \label{sec:opt}
\switchlanguage%
{%
  The overall system of this study is shown in \figref{figure:system}.
  Design parameters are defined, URDF models are automatically generated and evaluated, and black-box optimization is performed.
}%
{%
  本研究の全体システムを\figref{figure:system} に示す.
  まず設計パラメータを定義し, 次にURDFを自動生成, それを評価し, ブラックボックス最適化を行う.
}%

\subsection{Robot Design Parameters} \label{subsec:opt-params}
\switchlanguage%
{%
  In this study, the design parameters are varied between the simulations and the actual actuator modules.
  In the simulation, the optimization results are verified by using general design parameters, and in the actual robot, a body that can be configured with the actuator modules is designed and tested.
  All joints are single-axis joints that rotate in the pitch direction.
  As shown in the left figure of \figref{figure:system}, Link $i$ is connected to Joint $i$ ($0 \leq i \leq N_{joint}$, where $N_{joint}$ denotes the number of joints).
  Let $\bm{d}_i$ denote the direction vector of Link $i$ relative to the coordinates of Joint $i$, and $l_i$ denote the length of the link.
  Let $R_i$ be the rotation matrix of Joint $i$ with respect to the coordinates of Joint $i-1$.
  Note that there exists a fixed joint Joint $0$ at the origin of the world coordinates, which does not rotate.
  Joint 1 and Link 1 are the root of the robot body, and the origin of the robot is at the tip of Link 0.

  First, we describe the parameters for the general configuration in a simulation.
  The joint angle limit of each Joint $i$ is assumed to be -90 to 90 [deg].
  The rotation matrix $R_i$ of Joint $i$ has 12 discrete values in total, where the rotation axis of the joint faces each direction of $xyz$ and the joint center faces each of the four directions apart by 90 degrees.
  For $\bm{d}_i$, the direction vector of Link $i$, we prepare a total of six discrete values considering positive and negative values for each direction of $xyz$.
  In other words, at the initial position where all joint angles $\bm{\theta}$ are $\bm{0}$, all link directions are placed at right angles.
  The length of Link $i$, $l_i$, is assumed to be a continuous value between 0.1 and 0.6 [m].
  The links are constrained such that they do not overlap each other.

  Next, we describe the parameters for the actuator module configuration.
  Attachment 1 is attached to the actuator module at either end (two patterns), while Attachment 2 is attached at both ends (one pattern).
  For Attachment 1, the actuator module can be connected in a total of eight different ways.
  For Attachment 2, there are a total of six different link lengths and joint orientations, and then the actuator module can be connected in three different ways.
  Therefore, there are 26 ($= 8 + 6\times3$) discrete ways of connection in total.
  Note that $\bm{d}_0$ and $l_0$ in both general configuration and actuator module configuration are different from the above patterns.
  We prepare a three-dimensional position $\bm{a}$ which is the product of $\bm{d}_0$ and $l_0$, and set the range of its $xyz$ coordinates, $a_{\{x, y, z\}}$, as continuous values appropriate for each task.

  Based on these design parameters, URDF models are automatically generated.
  For the general configuration, all links are rectangular with 0.15 $\times$ 0.15 $\times$ $l_i$ [m].
  The mass and inertia of the links are calculated assuming that the density of the link is 1.0 g/cm$^3$.
  The generated URDF models are shown in \figref{figure:general-example}.
  These are examples when $N_{joint}=\{5, 10, 15\}$, and it can be seen that a variety of bodies are configured.
}%
{%
  設計パラメータを示す.
  本研究ではシミュレーションとアクチュエータモジュール実機でそのデザインパラメータを変化させている.
  シミュレーションにおいてはより一般的な形のパラメータを用いて最適化結果を検証し, 実機においては実際のアクチュエータモジュールで構成可能な身体を設計, 実験を行う.
  本研究では関節は全てピッチ方向に回転する1軸関節である.
  \figref{figure:system}の左図に示すように, Joint $i$に対してLink $i$が接続している ($0 \leq i \leq N_{joint}$, $N_{joint}$はジョイント数を表す).
  Joint $i$の座標系に対するLink $i$の方向ベクトルを $\bm{d}_i$, リンクの長さを$l_i$とする.
  また, Joint $i-1$の座標系に対するJoint $i$の回転行列を$R_i$とする.
  なお, 世界座標系原点の位置に回転しない固定関節Joint 0が存在し, その先にLink 0が存在する.
  Joint 1とLink 1からがロボットの本体であり, ロボットはLink 0の先端を原点とした位置から構成される.

  まずシミュレーションにおける一般的な形のパラメータについて述べる.
  それぞれのJoint $i$の関節角度限界を-90から90 [deg]とする.
  Joint $i$の回転行列である$R_i$は, 関節の回転軸が$xyz$それぞれの方向に向き, かつ関節中心が90度ずつ4方向それぞれに向く, 計12種類の離散値を用意する.
  Link $i$の方向ベクトルである$\bm{d}_i$は, $xyz$それぞれの方向についてプラスとマイナスを考慮した計6種類の離散値を用意する.
  つまり, 全ての関節角度$\bm{\theta}$が$\bm{0}$の初期位置において, 回転方向やリンク方向は全て直角に配置されている.
  リンク$i$の長さである$l_i$は, 0.1から0.6の間の連続値を用意する.
  なお, それらについて, リンク同士が重ならないような制約をかけている.

  次にアクチュエータモジュール実機におけるパラメータについて述べる.
  Attachment 1はアクチュエータモジュールに対して両端それぞれに取り付けることができ, Attachment 2は一通りの取り付け方しかない.
  Attachment 1に対しては, Actuator Moduleが4種類の方法で接続することが可能である.
  Attachment 2に対しては, リンクの長さと次の関節の向きが計6種類, その後3種類の方法で次のActuator Moduleを接続することができる.
  ゆえに, これらを合わせて26通りの離散的な接続方法を用意する.
  なお, general configurationとactuator module configurationのどちらも, $\bm{d}_0$と$l_0$についてはこれまでの選択肢とは異なる.
  $\bm{d}_0$と$l_0$をまとめた3次元位置$\bm{a}$を用意し, その$xyz$座標である$a_{\{x, y, z\}}$の範囲をタスクごとに適切に連続値として設定している.

  この設計パラメータをもとに, URDFを自動的に生成する.
  general configurationについて, リンクは全て長方形とし, は0.15 $\times$ 0.15 $\times$ $l_i$ [m]である直方体とする.
  リンクの密度を1.0 g/cm$^3$として, リンクの質量と慣性を計算している.
  生成されたURDFを\figref{figure:general-example}に示す.
  これは関節数が5, 10, 15の例であるが, 多様な身体が構成されていることがわかる.
}%

\begin{figure}[t]
  \centering
  \includegraphics[width=0.95\columnwidth]{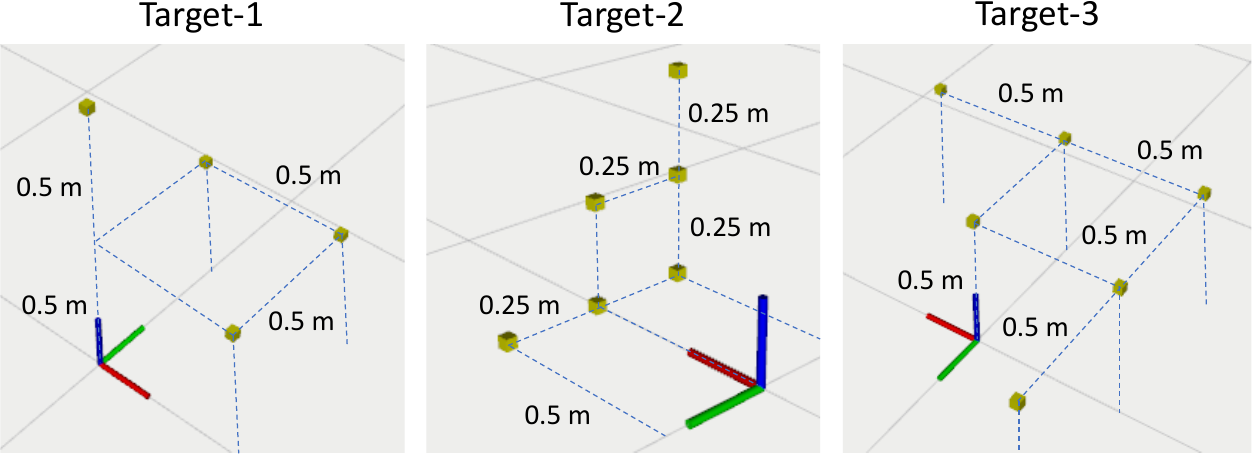}
  \caption{The target positions to be realized for general configuration.}
  \label{figure:sim-target}
\end{figure}

\begin{figure*}[t]
  \centering
  \includegraphics[width=1.8\columnwidth]{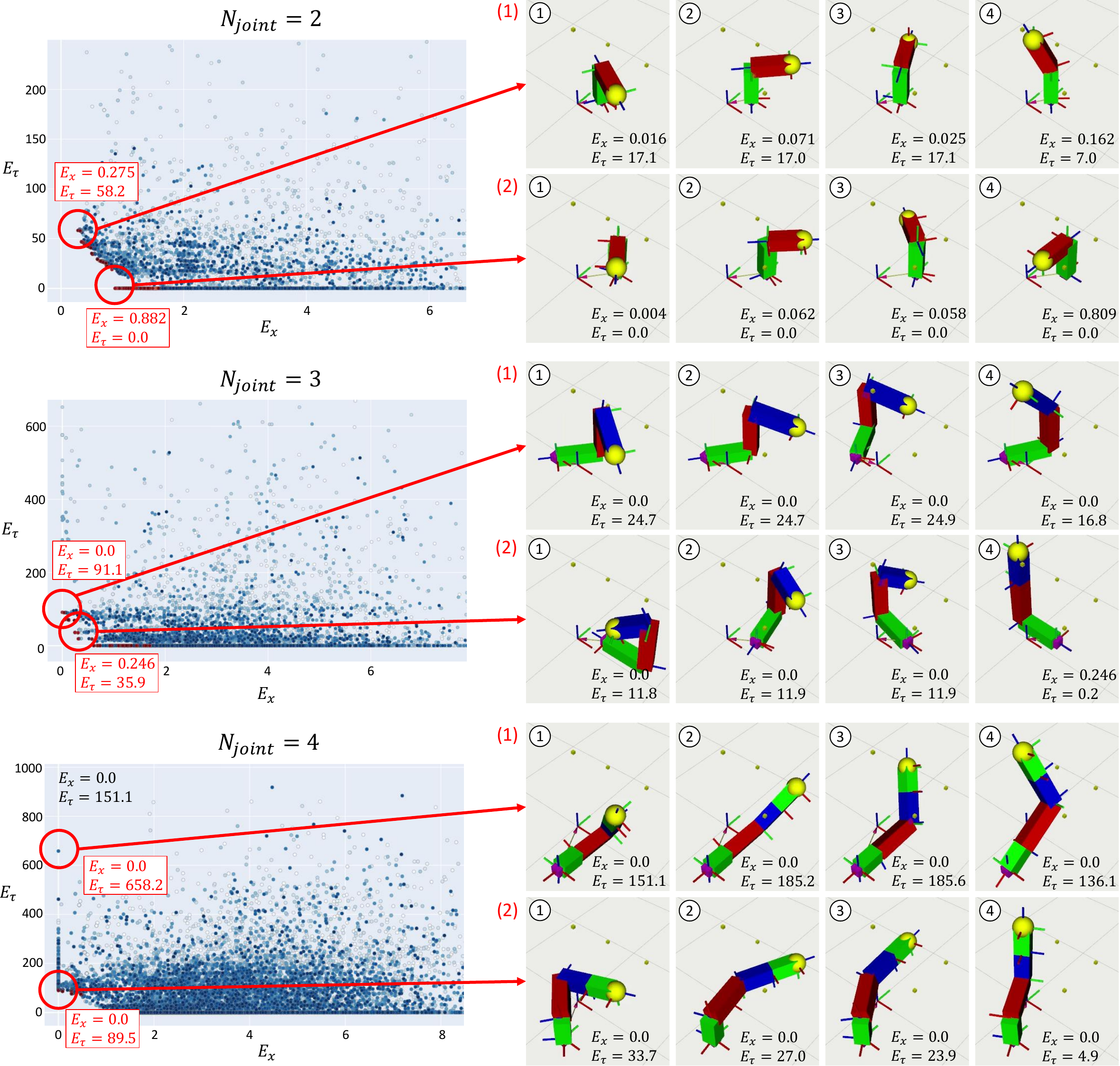}
  \caption{The optimization results for Target-1 with general configuration. $N_{joint}$ is changed to 2, 3, and 4, and two solutions are shown for each $N_{joint}$.}
  \label{figure:sim-1}
\end{figure*}

\begin{figure}[t]
  \centering
  \includegraphics[width=0.9\columnwidth]{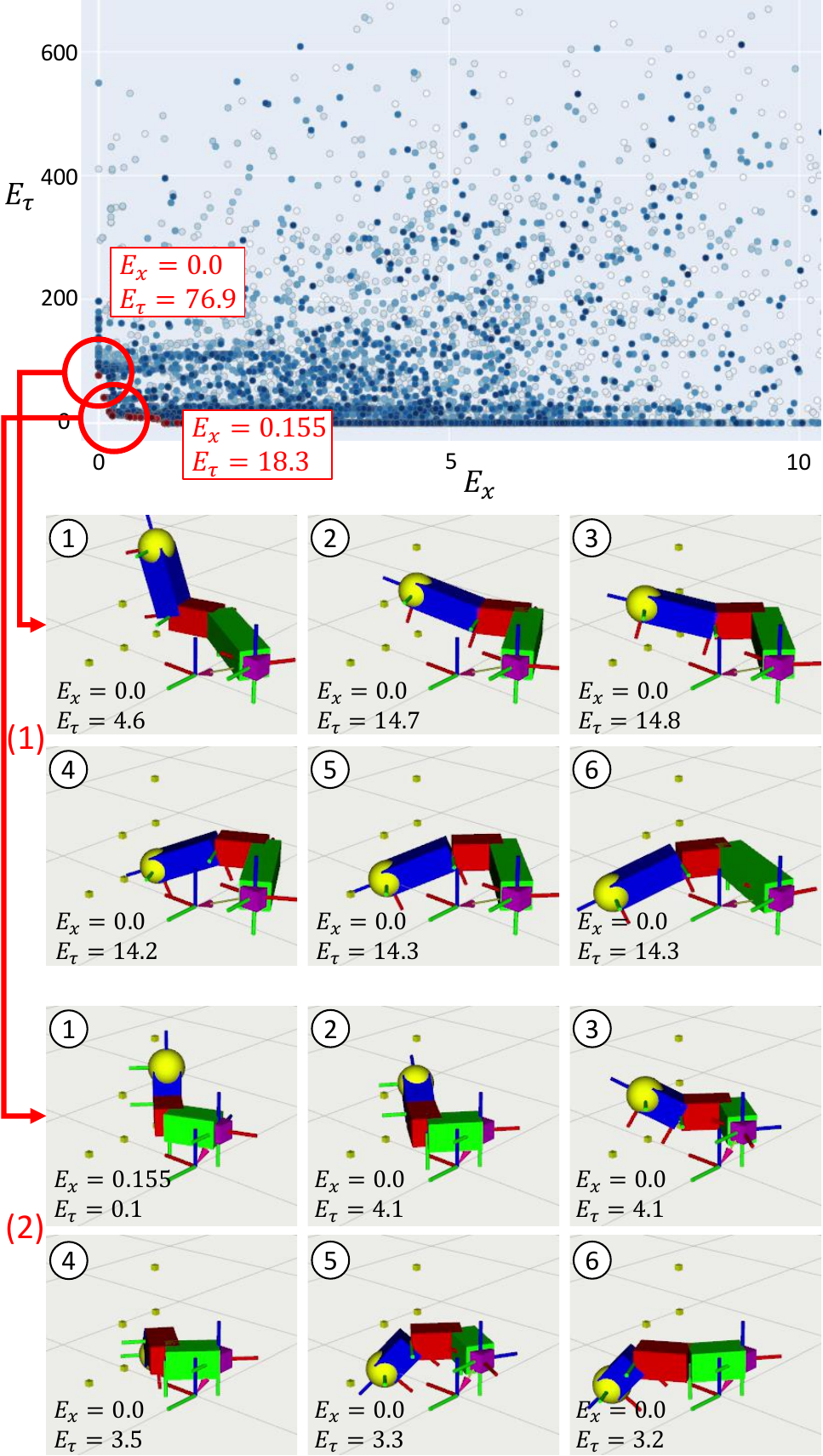}
  \caption{The optimization results for Target-2 with general configuration. Two Pareto front solutions are shown.}
  \label{figure:sim-2}
\end{figure}

\begin{figure}[t]
  \centering
  \includegraphics[width=0.9\columnwidth]{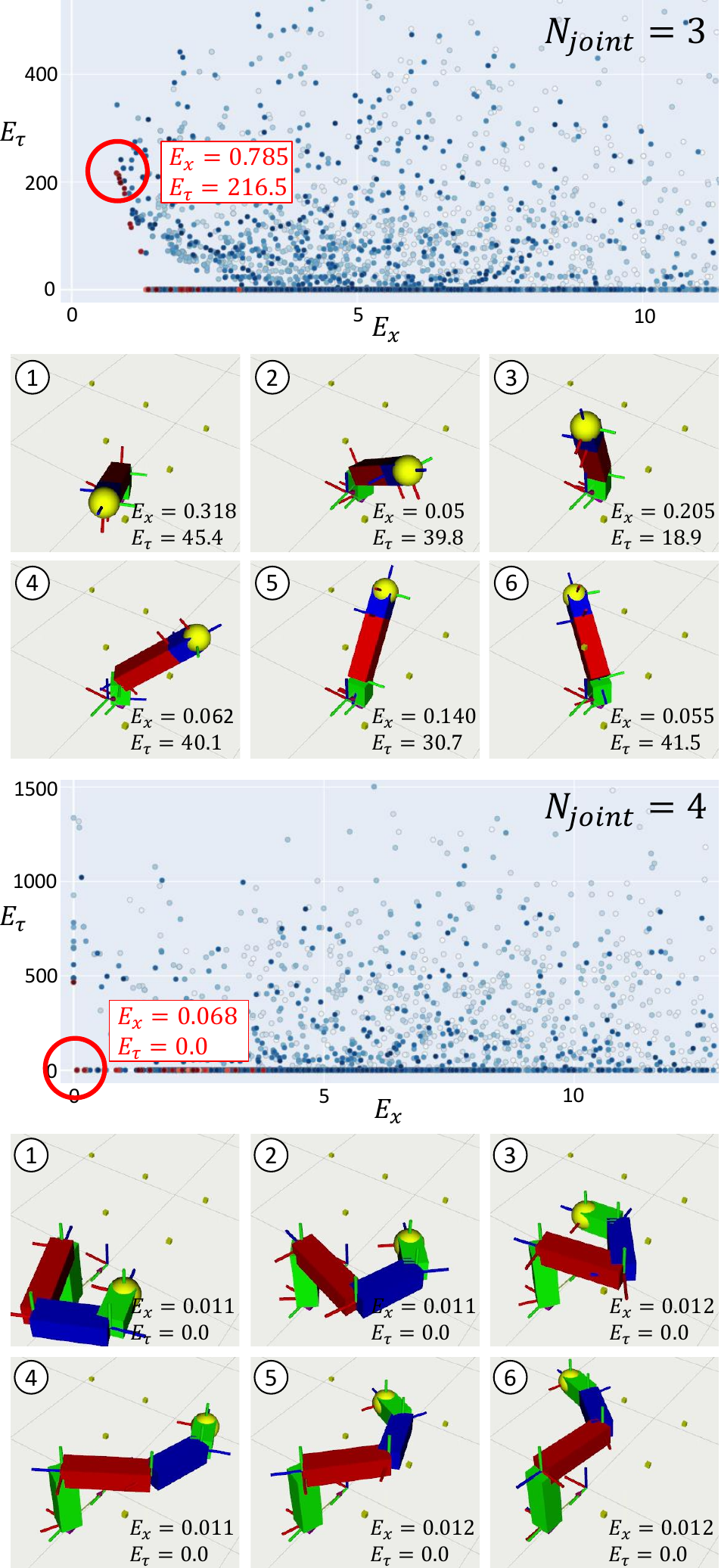}
  \caption{The optimization results for Target-3 with general configuration. $N_{joint}$ is changed to 3 and 4.}
  \label{figure:sim-3}
\end{figure}

\subsection{Black-Box Optimization of Design Parameters} \label{subsec:opt-eval}
\switchlanguage%
{%
  Although various forms of evaluation functions for body design optimization are possible, in this study, we adopt a relatively simple form due to the characteristics of the task.
  The evaluation values are the control error at the target coordinates and the joint torque value when reaching the target.
  First, the target position and posture $\{(\bm{x}^{ref}_1, R^{ref}_1), \cdots, (\bm{x}^{ref}_{N_{ref}}, R^{ref}_{N_{ref}})$ (where $N_{ref}$ is the number of target positions and postures) are given.
  Here, the following values $E_x$ and $E_{\tau}$ are calculated,
  \begin{align}
    \bm{x}_{i}, \bm{\tau}_{i} = \textrm{IK}(\bm{x}^{ref}_{i}, R^{ref}_{i})\\
    E_x = \sum^{N^{ref}}_{i} ||\bm{x}_{i}-\bm{x}^{ref}_{i}||_{2}\\
    E_{\tau} = \sum^{N^{ref}}_{i} ||\bm{\tau}_{i}||_{2}
  \end{align}
  where $\textrm{IK}$ is the inverse kinematics when the current URDF model is given, and $\bm{x}_{i}$ and $\bm{\tau}_{i}$ are the calculated end-effector position and joint torque.
  \cite{chan1995ik, sugiura2007ik} are used as the algorithm for the inverse kinematics.

  Generally, a single value $E=E_x + wE_{\tau}$ is calculated using a certain weight coefficient $w$, and optimization is performed based on this $E$.
  On the other hand, in such a case, it is difficult to optimize parameters appropriately because the solution varies depending on the adjustment of $w$.
  Moreover, since only one optimal solution can be obtained, it is impossible to create a process in which the user selects his/her preferred personalized body design based on the obtained solutions.
  Therefore, in this study, we perform a multi-objective optimization problem to minimize both $E_x$ and $E_{\tau}$ simultaneously, present several Pareto front solutions, and finally determine appropriate body parameters.
  We use Tree-Structured Parzen Estimator \cite{bergstra2011tpe} in Optuna \cite{akiba2019optuna} as a library of black-box optimization.
}%
{%
  身体設計最適化の評価関数は多様な形式が可能であるが, 本研究ではタスクの特性上比較的シンプルな形を取っている.
  評価値として扱うのは, 指令したポイントへの到達精度と, その際の関節トルク値である.
  まず, 指令位置と指令姿勢$\{(\bm{x}^{ref}_1, R^{ref}_1), \cdots, (\bm{x}^{ref}_{N_{ref}}, R^{ref}_{N_{ref}})$ ($N_{ref}$は指令位置姿勢の数を表す)を与える.
  ここで, 以下の値$E_x$と$E_{\tau}$を計算する.
  \begin{align}
    \bm{x}_{i}, \bm{\tau}_{i} = \textrm{IK}(\bm{x}^{ref}_{i}, R^{ref}_{i})\\
    E_x = \sum^{N^{ref}}_{i} ||\bm{x}_{i}-\bm{x}^{ref}_{i}||_{2}\\
    E_{\tau} = \sum^{N^{ref}}_{i} ||\bm{\tau}_{i}||_{2}
  \end{align}
  ここで, $\textrm{IK}$は現在のURDFを用いた際の逆運動学, $\bm{x}_{i}$と$\bm{\tau}_{i}$は計算された手先位置とその際の関節トルクである.
  なお, 逆運動のアルゴリズムは\cite{chan1995ik, sugiura2007ik}におけるアルゴリズムを用いている.

  これらは一般的にはある重み係数$w$を用いて単一の値$E_x + wE_{\tau}$を計算し, これに基づいて最適化を行う.
  一方その場合, $w$の調整次第で解が異なるため適切な最適化が難しい.
  また, 一つの最適解しか得られないため, 得られた解を見てユーザが好みのパーソナライズされた身体形態を選ぶ過程を作ることができない.
  そこで本研究では, この$E_x$と$E_{\tau}$の両者を同時に最小化する多目的最適化問題を行い, いくつかのパレート解を提示し最終的に適切な身体パラメータを決定する.
  なお, 本研究ではBlack-Box OptimizationのライブラリとしてOptuna \cite{akiba2019optuna}のTree-Structured Parzen Estimator \cite{bergstra2011tpe}を用いている.
  $E_x$と$E_{\tau}$の最小化を設定し, 試行回数はタスクごとに異なる.
}%

\section{Experiments} \label{sec:experiment}
\switchlanguage%
{%
  Various experiments are first conducted in the general configuration to demonstrate the effectiveness of this study.
  Next, experiments are conducted on actual robots with the actuator module configuration to demonstrate the practical applications of this study.
}%
{%
  本研究では, まずgeneral configurationにおいて多様な実験を行い, 本研究の有効性を示す.
  その後actuator module configurationにおいて実機実験を行い, 実応用例を示す.
}%

\subsection{Automatic Design Optimization for General Configuration} \label{subsec:basic-exp}
\switchlanguage%
{%
  Target-1, Target-2, and Target-3 shown in \figref{figure:sim-target} are prepared as target positions (target postures are not given).
  $a_{\{x, y, z\}}$ is specified in the range of [-1.0, 1.0] [m], but $a_{z}$ is set to [-0.1, 0.1] for Target-1, $a_{x}$ is set to [-1.0, 0.0] for Target-2, and $a_{z}$ is set to [-1.0, 0.0] for Target-3.

  For Target-1, the sampling results when $N_{joint}=\{2, 3, 4\}$ are shown in the left figures of \figref{figure:sim-1}, and the inverse kinematics results for some of the solutions are shown in the right figure of \figref{figure:sim-1}.
  Here, the red dots in the graphs are Pareto front solutions.
  For $N_{joint}=2$, solution (1) is a design in which $E_{x}$ is small but $E_{\tau}$ is large, and solution (2) is a design in which $E_{x}$ is large but $E_{\tau}$ is zero.
  With $N_{joint}=2$, it is difficult to satisfy all the target positions, and $E_{x}$ cannot be zero.
  In (1), $E_{x}$ is reduced as much as possible by installing a yaw-axis joint in the first axis and a pitch-axis joint in the second axis, although the required torque increases.
  On the other hand, in (2), both the first and second axes are yaw-axis joints, so that the required torque is always zero.
  However, due to the two-dimensional motion of yaw-axis joints, only 3 out of 4 target positions can be supported, and $E_{x}$ is significantly increased.
  Next, for $N_{joint}=3$, solution (1) is a design in which $E_{x}$ is zero, and solution (2) is a design in which $E_{\tau}$ is reduced as much as possible while maintaining accuracy.
  By increasing the number of joints, (1) is able to realize all target positions, while $E_{\tau}$ is higher than that of (1) with $N_{joint}=2$.
  On the other hand, in (2), the length of the final link is shortened to prevent the increase of $E_{\tau}$.
  Finally, for $N_{joint}=4$, solution (1) is a design with the largest $E_{\tau}$ among $E_{x}=0$ apart from the Pareto front, and solution (2) is a design with the smallest $E_{\tau}$ among $E_{x}=0$.
  (2) with $N_{joint}=4$ is almost the same as (1) with $N_{joint}=3$, which means that $N_{joint}=3$ is sufficient for Target-1.
  The above mentioned solutions reduced the required torque by using yaw-axis joints for the root joints, but (1) with $N_{joint}=4$ uses a pitch-axis joint for the root joint, resulting in a large $E_{\tau}$.
  It can be seen that the required torque varies greatly depending on the joint arrangement.

  For Target-2, the sampling result when $N_{joint}=3$ is shown in the upper figure of \figref{figure:sim-2}, and the inverse kinematics results for two of the Pareto front solutions are shown in the lower figure of \figref{figure:sim-2}.
  Solution (1) is a design that satisfies $E_{x}=0$ for all the target positions but has a large $E_{\tau}$, while solution (2) is a design with reduced $E_{\tau}$ but with slightly lower accuracy.
  In both (1) and (2), the root joint is a yaw-axis joint that is not subject to gravity force, and only the tip joint is a pitch-axis joint.
  The link length of (2) is shorter than that of (1), which reduces the required torque significantly while somewhat decreasing the accuracy.

  For Target-3, the sampling results when $N_{joint}=\{3, 4\}$ and the inverse kinematics results for one Pareto front solution for each are shown in \figref{figure:sim-3}.
  The solution with the smallest $E_{x}$ is shown for $N_{joint}=3$, but $E_{x}$ is not zero and $E_{\tau}$ is large.
  The target coordinates of Target-3 are similar to those of Target-2, and $N_{joint}=3$ seems to be able to achieve $E_{x}=0$, but the widths of the target positions are wider than that of Target-2, and the link length is insufficient to achieve $E_{x}=0$.
  On the other hand, for $N_{joint}=4$, a design where $E_{x}$ is almost zero and $E_{\tau}$ is also zero is generated.
  The target positions of Target-3 are aligned on the plane where $z=0.5$, and thus a so-called SCARA type robot is generated.
}%
{%
  指令位置として\figref{figure:sim-target}に示すTarget-1, Target-2, Target-3を用意した(指令姿勢は与えていない).
  なお, 基本的に$a_{\{x, y, z\}}$は[-1.0, 1.0] [m]の範囲を指定するが, Target-1については$a_{z}$を[-0.1, 0.1], Target-2については$a_{x}$を[-1.0, 0.0], Target-3については$a_{z}$を[-1.0, 0.0]に変更している.

  Target-1について, $N_{joint}=\{2, 3, 4\}$としたときのサンプリング結果を\figref{figure:sim-1}の左図に, そのうちのいくつかの解における逆運動学の結果を\figref{figure:sim-1}の右図に示す.
  ここで, グラフにおける赤い点がPareto front解である.
  まず, $N_{joint}=2$における(1)の解は$E_{x}$が0.275と小さいが$E_{\tau}$が大きな設計, (2)の解は$E_{x}$が0.882と大きいが$E_{\tau}$が0な設計である.
  $N_{joint}=2$ではまだ指令位置を全て満たす事が難しく, $E_{x}$は0となることが出来ていない.
  (1)では1軸目にyaw軸関節, 2軸目にpitch軸関節を設けることで, 必要トルクは上昇するが, $E_{x}$を最大まで削減している.
  一方(2)では, 1軸目, 2軸目をともにyaw軸関節とすることによって, 常に必要なトルクを0としている.
  そのため, 4つの指令位置のうち3つにしか対応できず, $E_{x}$が大きく上昇している.
  次に, $N_{joint}=3$における(1)の解は$E_{x}$が0となる設計であり, (2)の解は$E_{\tau}$をなるべく下げつつ精度を保った設計である.
  関節数を増やすことで, (1)は全ての指令位置を実現することが可能となっている一方, $E_{\tau}$は$N_{joint}=2$の(1)よりも上昇している.
  これに対して, (2)は最終リンクの長さを短くすることで, $E_{\tau}$の上昇を防いだ形である.
  最後に, $N_{joint}=4$における(1)の解はPareto frontから外れた$E_{x}=0$の中で最も$E_{\tau}$が大きい設計, (2)の解は$E_{x}=0$の中で最も$E_{\tau}$が小さな設計である.
  (2)については$N_{joint}=3$の(1)とほとんど変わらず, つまりTarget-1については$N_{joint}=3$で十分であることがわかる.
  これまでの解は根本の関節をyaw軸関節とすることで必要トルクを削減していたが, (1)の解は根本からpitch軸関節を使っているため, $E_{\tau}$が非常に大きい.
  関節の配置一つで必要トルクが大きく変化することが見て取れる.

  Target-2について, $N_{joint}=3$としたときのサンプリング結果を\figref{figure:sim-2}の上図に, そのうち2つのPareto front解における逆運動学の結果を\figref{figure:sim-2}の下図に示す.
  ここで, (1)の解は$E_{x}=0$と指令位置を全て満たすが$E_{\tau}$が大きな設計, (2)の解は精度は多少落とすが$E_{\tau}$を削減した形の設計である.
  (1)と(2)のどちらも根本の関節を重力の力を受けないyaw軸関節にし, 先端のみをpitch軸関節としている.
  (2)については, (1)よりもリンク長を短くすることで, 精度を多少落としながらも, 必要トルクを大幅に削減している.

  Target-3について, $N_{joint}=\{3, 4\}$としたときのサンプリング結果と, それぞれについて1つのPareto front解における逆運動学の結果を\figref{figure:sim-3}に示す.
  $N_{joint}=3$の際に最も$E_{x}$が小さい解を示しているが, $E_{x}$は0でないうえに$E_{\tau}$は非常に大きい.
  Target-3はTarget-2とよく似ており, $N_{joint}=3$でも$E_{x}=0$を達成できそうであるが, Target-2よりも指令位置の幅が広く, リンク長が足りないためこのような結果となっている.
  一方で, $N_{joint}=4$では$E_{x}$がほとんど0かつ$E_{\tau}$も0である設計が生成されている.
  Target-3は指令位置が$z=0.5$となる平面上に並ぶため, いわゆるスカラ型のロボットが生成されていることがわかる.
}%

\begin{figure}[t]
  \centering
  \includegraphics[width=0.8\columnwidth]{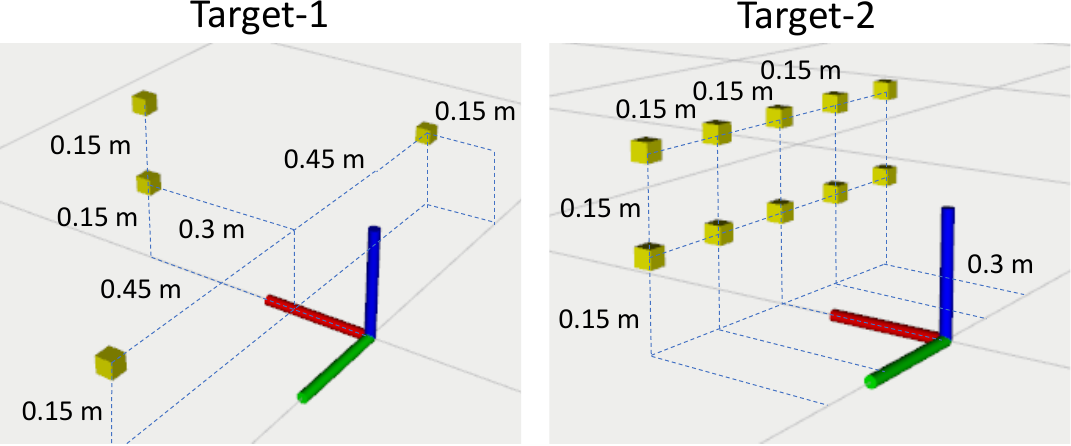}
  \caption{The target positions to be realized for actuator module configuration.}
  \label{figure:act-target}
\end{figure}

\begin{figure*}[t]
  \centering
  \includegraphics[width=1.9\columnwidth]{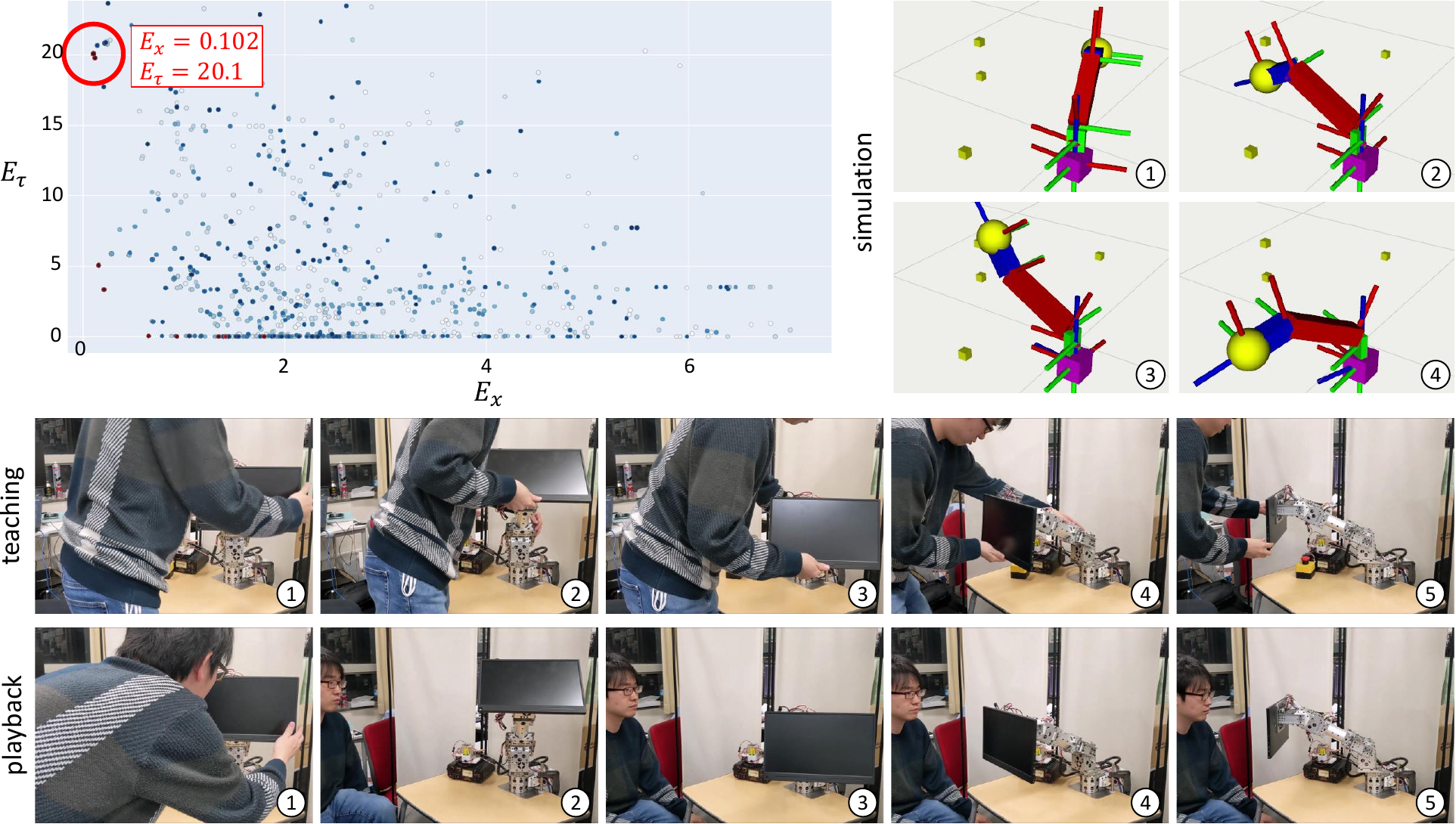}
  \caption{The optimization results for Target-1 with actuator module configuration. One Pareto front solution is shown in simulation. The lower figures show the teaching and playback motions of the actual optimized modular robot for monitor movement.}
  \label{figure:act-1}
\end{figure*}

\begin{figure*}[t]
  \centering
  \includegraphics[width=1.9\columnwidth]{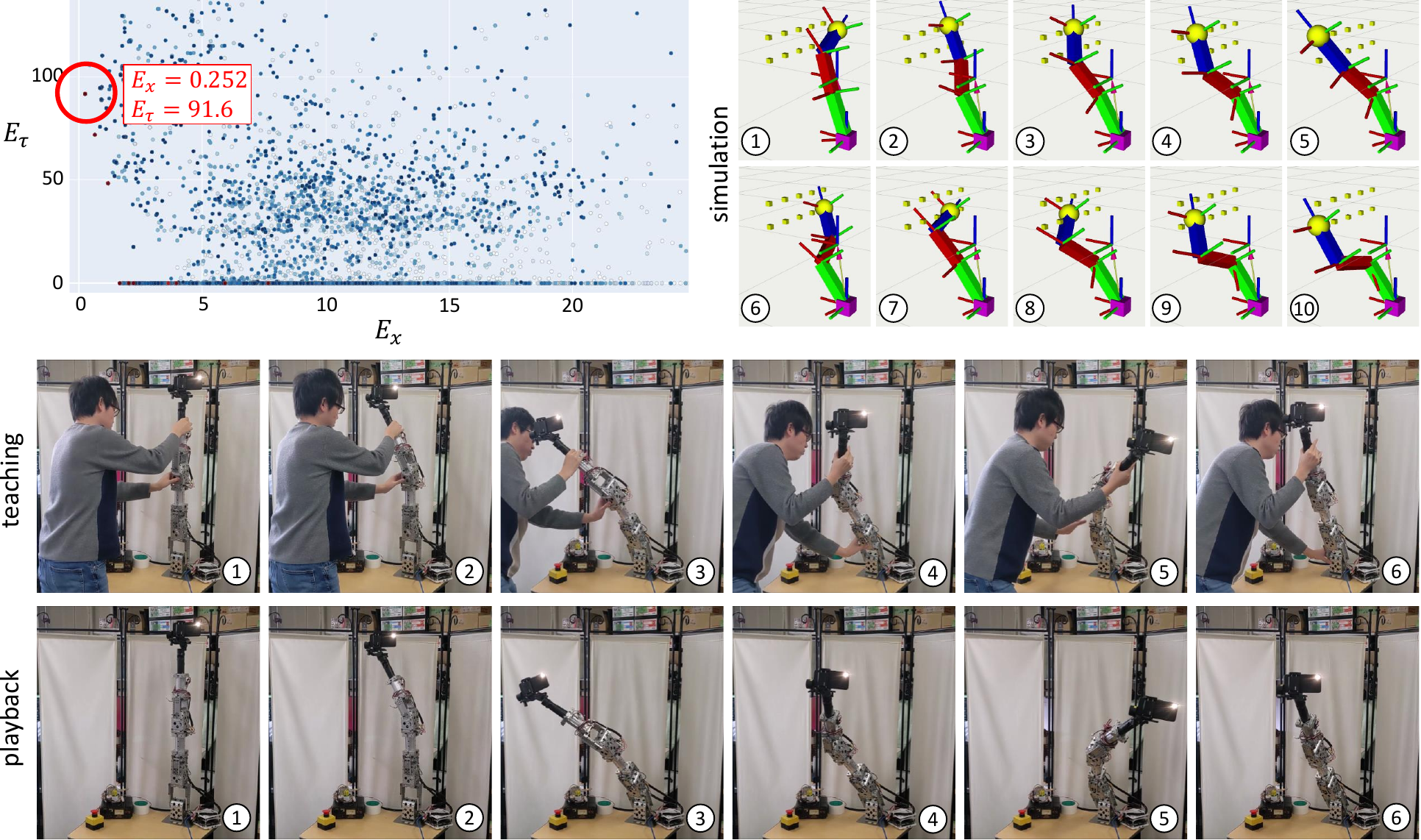}
  \caption{The optimization results for Target-2 with actuator module configuration. One Pareto front solution is shown in simulation. The lower figures show the teaching and playback motions of the actual optimized modular robot for a lighting operation.}
  \label{figure:act-2}
\end{figure*}

\subsection{Automatic Design Optimization for Actuator Module Configuration} \label{subsec:advanced-exp}
\switchlanguage%
{%
  Target-1 and Target-2 shown in \figref{figure:act-target} are prepared as the target positions.
  Note that $a_{\{x, z\}}$ and $a_y$ are specified in the range [-1.0, 0.0] [m] and [-1.0, 1.0] [m], respectively.

  For Target-1, the sampling results and inverse kinematics results for one Pareto front solution are shown in the upper figures of \figref{figure:act-1}.
  Since the target positions are distributed over a wide range, the range is covered by using a yaw-axis joint at the root of the body and two pitch-axis joints next to it.
  The actual motions of the modular robot built with the automatically designed parameters are shown in \figref{figure:act-1}.
  As an example, the task of moving a monitor over a wide range is conducted by attaching a monitor to the end of the arm.
  By using the lock-release mechanism of the module, the robot is operated by direct teaching in the released state and successfully moves a heavy monitor in the locked state by playing back the taught motion.
  Since all modules are equipped with IMUs, it is possible to detect vibrations from humans to the monitor and replay the taught motion automatically.

  For Target-2, the sampling results and the inverse kinematics results for one Pareto front solution are shown in the upper figures of \figref{figure:act-2}.
  In order to satisfy a large number of target positions in the upper front of the robot, a pitch-axis joint is used at the root of the body and two yaw-axis joints are used next to it.
  The actual motions of the modular robot built with the automatically designed parameters are shown in \figref{figure:act-2}.
  As an example, the task of lighting operation is conducted by attaching a light to the end of the arm.
  By using the lock-release mechanism of the module, the robot is operated by direct teaching in the released state and successfully operates the light in the locked state.
  In a similar manner, it is also possible for the robot to operate a camera to take pictures or serve food.
}%
{%
  指令位置として\figref{figure:act-target}に示すTarget-1とTarget-2を用意した.
  なお, それぞれ$a_{\{x, z\}}$は[-1.0, 0.0] [m], $a_y$は[-1.0, 1.0] [m]の範囲を指定する.

  Target-1について, サンプリング結果とパレートフロント解のうちの一つの解における逆運動学の結果を\figref{figure:act-1}の上図に示す.
  指令位置が広い範囲に分布しているため, 根本をyaw軸関節として, その先をpitch関節2つとすることで広い動作範囲をカバーしている.
  自動設計されたパラメータにより実際に構築したモジュラーロボットの動作を\figref{figure:act-1}の下図に示す.
  例として広い範囲のモニタ移動を考え, アーム先端にモニタを取り付けた.
  モジュールのロック・フリー機構を用いることで, フリー状態で動作を教示し, ロック状態で重いモニタを移動させることに成功している.
  全モジュールにIMUを搭載するため, 人間からのモニタへの振動等を検知して教示動作を再生することが可能である.

  Target-2について, サンプリング結果とパレートフロント解のうちの一つの解における逆運動学の結果を\figref{figure:act-2}の上図に示す.
  前方の多数の指令位置を満たすために, 根本をpitch軸関節として, その先をyaw関節2つとしている.
  自動設計されたパラメータにより実際に構築したモジュラーロボットの動作を\figref{figure:act-2}の下図に示す.
  例としてライティング操作を考え, アーム先端にライトを取り付けた.
  モジュールのロック・フリー機構を用いることで, フリー状態で動作を教示し, ロック状態でライトを操作させることに成功している.
  同様の形でカメラによる撮影や配膳等を行うことも可能である.
}%

\section{Discussion} \label{sec:discussion}
\switchlanguage%
{%
  The results obtained are summarized below.
  In this study, several target coordinates that should be realized by a robot are given, body structures with minimum control error and minimum torque are generated by simulations, and one of them is selected to construct an actual modular robot.
  From the simulation experiments, it is found that the obtained Pareto front solutions have various reasonable joint arrangements and link lengths.
  In order to minimize the torque, the yaw-axis joints are arranged at the root of the body to avoid the effect of gravity, and the link lengths are shortened as much as possible.
  The optimal number of joints can be obtained by running the optimization while changing the number of joints.
  Depending on the target coordinates, a general configuration such as a SCARA type robot can also be obtained.
  Next, from the actual robot experiments, the obtained optimal solution is actually constructed and the desired performance is successfully obtained.
  In particular, the actuator module with a lock-release mechanism enables the robot to carry heavy objects without back drive and to be operated by direct teaching.
  The actual robots have successfully moved monitors and performed lighting operations.
  Similar operations such as food delivery, tool delivery, and camera operation are also possible.

  The use of Tree-Structured Parzen Estimator is versatile, and it can solve various problems of handling continuous and discrete parameters.
  Therefore, it is possible to solve a very wide range of optimization problems including kinematic branching and body dynamics in a unified manner by parameterizing parent links, gear ratios, and so on.
  On the other hand, the number of parameters is fixed due to the nature of black-box optimization.
  Of course, it is possible to define a large number of parameters and add another variable that expresses the range of the parameters to be used, but at present, each optimization is performed while changing the numbers of joints.
  Although we did not handle complicated body parameters or evaluation functions in this study due to the nature of daily assistive robots, we would like to increase the degrees of freedom and handle more complicated parameters in the future.
  Also, an actuator module with a lock-release mechanism was used to carry heavy objects and be operated by direct teaching, but these characteristics are not directly incorporated into the optimization.
  In the future, we would like to study which part of the module should have the lock-release mechanism in combination with general actuator modules.
  Another important issue to be addressed is how to construct a novel body configuration combined with linear joints and tendon-driven actuators through black-box optimization.
}%
{%
  得られた結果についてまとめる.
  本研究ではロボットが実現すべき多数の指令座標を与え, これを満たしつつトルク最小な身体構造をシミュレーションから生成, そのうち一つを選んでモジュラーロボット実機を構築した.
  シミュレーション実験から, 得られたパレートフロント解は異なる多様な関節配置やリンク長を有しており, それらはリーズナブルなものであることがわかった.
  トルクを最小化するために, 重力を受けないよう根本にyaw軸関節を配置したり, なるべくリンク長を短くしたりする様子が見て取れた.
  また, 関節数を変化させながら最適化を実行することで, 最適な関節数を得ることができる.
  指令座標によっては, スカラ型ロボットのような一般的な構成も見ることができる.
  次に実機実験から, 得られた最適解を実際に構築し, 所望の性能を引き出すことに成功した.
  特に, ロック・フリー機構を有するアクチュエータモジュールによって, 逆駆動せずに重量物を運ぶことと, 人間によるダイレクトティーチングが可能なことを両立できている.
  実際にモニタ移動とライティング操作を行ったが, 同様の形で配膳や道具受け渡し, カメラ操作等が可能であると考えられる.

  本手法におけるTree-Structured Parzen Estimatorの利用は汎用的であり, 連続値や離散値の混合問題を自由に解くことができる.
  ゆえに, 親リンクの指定やギア比等を含めることで, 枝分かれやダイナミクスの考慮等, 非常に広い範囲の最適化問題を統一的に解くことが可能である.
  その一方で, ブラックボックス最適化という性質の都合上, パラメータ数が固定である点は問題である.
  もちろん, 多めにパラメータを定義しておき, どこまでを用いるかを表現するパラメータを追加することもできるが, 現状では関節数を変えながらそれぞれ最適化を実行している.
  本研究は日常生活支援ロボットの設計ということであまり複雑な身体や評価関数は与えなかったが, 今後より自由度を増やし, 複雑なパラメータを扱っていきたい.
  また, 本研究ではロック・リリース機構を有するアクチュエータモジュールにより重量物可搬性能と教示性能を両立したが, これを直接最適化に組み込んでいるわけではない.
  今後, 一般的なアクチュエータモジュールと組み合わせ, どの部分にロック・リリース機構を持たせるべきか等についても検討していきたい.
  その他, 直動関節や腱駆動機構と合わせた新しい身体構成を最適化により如何に構築していくかが, 今後の重要な課題である.
}%

\section{CONCLUSION} \label{sec:conclusion}
\switchlanguage%
{%
  In this study, we have developed a modular robot for daily life support that is personalized to each user and task.
  The actuator module has a lock-release mechanism and allows various configurations of links and joints, which enables manipulation of heavy objects and direct teaching.
  By changing the orientation of the two types of attachments and the length of the links, a variety of bodies can be constructed.
  By optimizing these body design parameters with Tree-structured Parzen Estimator, a type of black-box optimization, a body that can achieve the desired motion with a small required torque can be constructed automatically.
  The optimized body was actually constructed and successfully used to operate monitors and lights.
  In the future, we would like to expand the range of the actuator design, target motions, and evaluation functions, toward developing robots that can adapt to various environments.
}%
{%
  本研究では, ユーザやタスクにパーソナライズされた日常生活支援・介護支援等に向けたロボットを目指したモジュラーロボットの開発を行った.
  ロック・フリー機構を有し, 多様なリンク・関節構成が可能なアクチュエータモジュールにより, 重量物体操作や教示が可能である.
  2種類のモジュールとそのアタッチメントの向き, リンク長を変化させることで, 多様な身体を構築可能である.
  この身体パラメータをブラックボックス最適化であるTPEにより最適化することで, 小さな必要トルクで所望の動作を実現可能な身体を自動構築できる.
  最適化された身体を実際に構築し, モニタやライトの動き, 運搬等に成功した.
  今後, アクチュエータ設計や目的動作, 評価関数の幅を増やし, 多様な環境に適応したロボットを開発していきたい.
}%

\section*{Acknowledgement}
This work was partially supported by JST Moonshot R\&D under Grant Number JPMJMS2033.

{
  \bibliographystyle{IEEEtran}
  \bibliography{main}
}

\end{document}